\documentclass{article}

\usepackage{microtype}
\usepackage{graphicx}
\usepackage{subfigure}
\usepackage{booktabs} 
\usepackage[table]{xcolor}
\usepackage{multirow}

\usepackage{hyperref}

 \usepackage{amsmath}
 \usepackage{amsfonts}
\newcommand{\Var}{\mathrm{Var}}
\newtheorem{Theorem}{Theorem}
\newtheorem{Lemma}[Theorem]{Lemma}

\usepackage[accepted]{mlsys2025}

\begin{document}

\twocolumn[
\mlsystitle{GradientSpace: Unsupervised Data Clustering for Improved Instruction Tuning}



\mlsyssetsymbol{equal}{*}

\begin{mlsysauthorlist}
\mlsysauthor{Shrihari Sridharan}{to}
\mlsysauthor{Deepak Ravikumar}{to}
\mlsysauthor{Anand Raghunathan}{to}
\mlsysauthor{Kaushik Roy}{to}
\end{mlsysauthorlist}


\mlsysaffiliation{to}{School of Electrical and Computer Engineering, Purdue University}

\mlsyscorrespondingauthor{Shrihari Sridharan}{sridhar4@purdue.edu}

\mlsyskeywords{Machine Learning, MLSys}

\vskip 0.3in

\begin{abstract}
Instruction tuning is one of the key steps required for adapting large language models (LLMs) to a broad spectrum of downstream applications. However, this procedure is difficult because real-world datasets are rarely homogeneous; they consist of a mixture of diverse information, causing gradient interference, where conflicting gradients pull the model in opposing directions, degrading performance. A common strategy to mitigate this issue is to group data based on semantic or embedding similarity. However, this fails to capture how data influences model parameters during learning. While recent works have attempted to cluster gradients directly, they randomly project gradients into lower dimensions to manage memory, which leads to accuracy loss. Moreover, these methods rely on expert ensembles which necessitates multiple inference passes and expensive on-the-fly gradient computations during inference. To address these limitations, we propose \textsc{GradientSpace}, a framework that clusters samples directly in full-dimensional gradient space. We introduce an online SVD-based algorithm that operates on LoRA gradients to identify latent skills without the infeasible cost of storing all sample gradients. Each cluster is used to train a specialized LoRA expert along with a lightweight router trained to select the best expert during inference. We show that routing to a single, appropriate expert outperforms expert ensembles used in prior work, while significantly reducing inference latency. Our experiments across mathematical reasoning, code generation, finance, and creative writing tasks demonstrate that \textsc{GradientSpace} leads to coherent expert specialization and consistent accuracy gains over state-of-the-art clustering methods and finetuning techniques.

\end{abstract}
]



\printAffiliationsAndNotice{}  

\section{Introduction}
\label{intro}
Large language models (LLMs) like GPT~\cite{openai2024}, Llama~\cite{llama3}, and DeepSeek~\cite{deepseekai} have become integral components in modern AI systems such as conversational agents and chatbots due to their ability to generate coherent text that closely mirrors natural human communication. One of the key reasons for this success is instruction tuning, a procedure where pretrained models are finetuned on datasets that consist of human-written instructions paired with responses~\cite{itsurvey1, itsurvey2}. 

Although instruction tuning has shown good efficacy, the datasets are usually curated from a single source or belong to a specific domain, which limits their ability to generalize across diverse real-world data distributions~\cite{limitsit,itsurvey2}. For example, when building an internal knowledge assistant within an organization, the underlying training data may include a mixture of documents, emails, support tickets, and wiki pages. Recent works~\cite{datadiv, llavamole, mixskills} have demonstrated that when instructions are mixed naively from different datasets, the performance of the model degrades. This is primarily due to a phenomenon called gradient interference~\cite{pcgrad, recon}, where examples corresponding to distinct tasks or domains push model parameters in conflicting directions. Learning from  diverse sources is challenging because the model must reconcile competing learning signals, leading to negative transfer and reduced overall performance.

To address this challenge, previous efforts have explored strategies to mitigate gradient interference. One such approach groups training examples by semantic similarity, assuming that semantically related inputs would influence compatible parameter updates~\cite{commonit, car}. However, input similarity is not a reliable proxy for how the model learns these inputs. Two semantically similar inputs may produce opposing gradients, while unrelated examples can reinforce each other if their updates are aligned. As shown in Figure~\ref{fig:motiv}, gradient and semantic similarities show little correlation, highlighting the need for organizing data based on learning dynamics. A different line of research, known as gradient surgery, modifies gradients during training to remove conflicts. Methods such as ~\cite{pcgrad} and ~\cite{cagrad} project gradients from different tasks onto non-conflicting subspaces to reduce interference. However, these approaches rely on explicit task labels to project gradients, which limits their applicability to real-world instruction tuning.

\begin{figure}[htb]
  \includegraphics[width=\columnwidth]{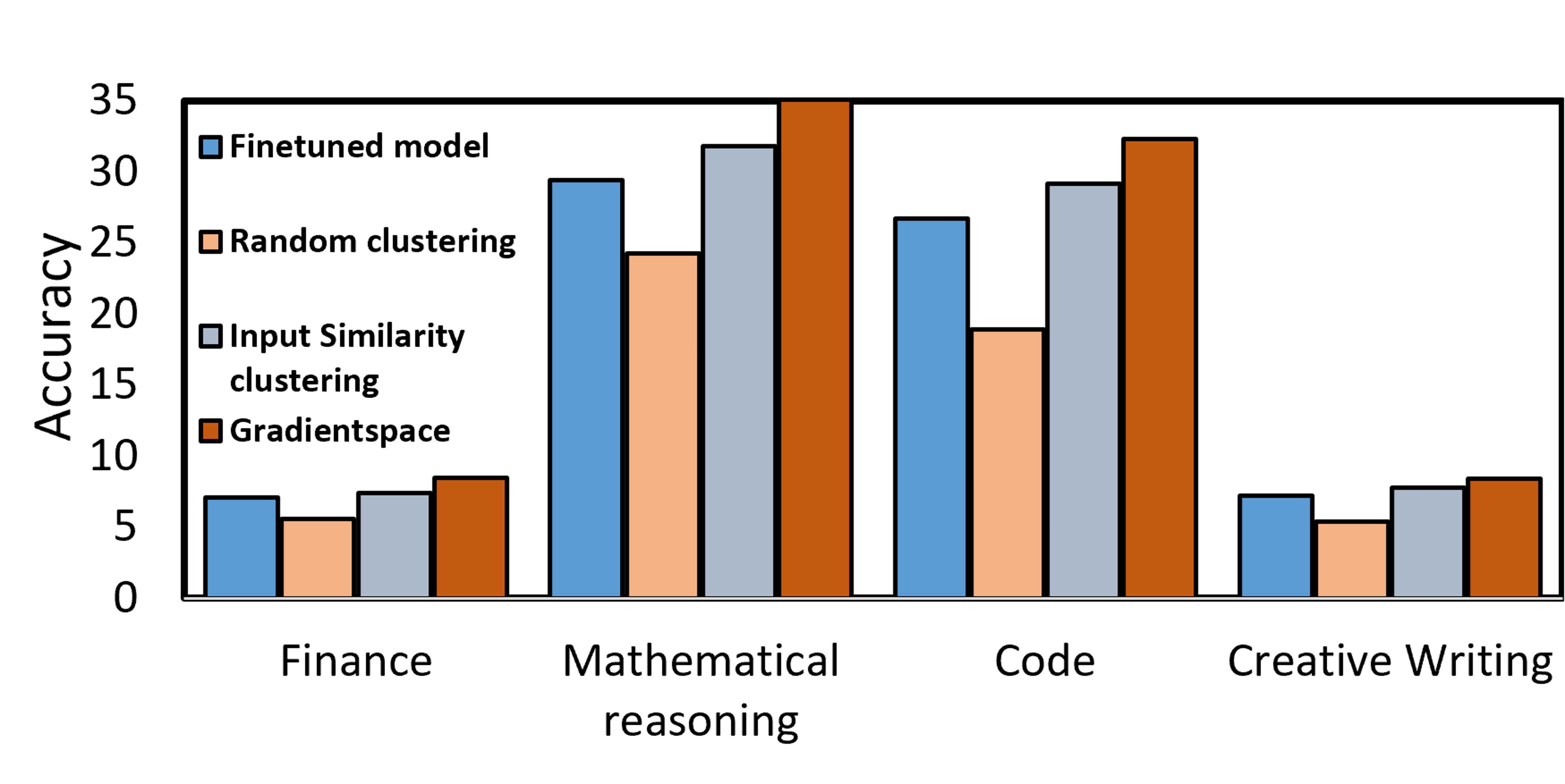}
  \caption{\textsc{GradientSpace} accuracy compared to other methods on different benchmarks. }
  \label{fig:modelacc}
\end{figure}

This motivates a fundamental question: \textit{``How should finetuning data be partitioned effectively when task boundaries are unknown?"} To address this, we introduce \textsc{GradientSpace}, a framework that clusters data based on gradient similarity and groups samples that produce positively correlated parameter updates. \textsc{GradientSpace} naturally identifies latent “skills” within unlabeled data by clustering examples that produce positively aligned gradient updates. Theoretically, this is equivalent to the covariance term in the gradient variance decomposition, showing that grouping examples with high positive covariance reduces gradient interference and improves Stochastic Gradient Descent (SGD) convergence (more details in Section ~\ref{sec:theory}). Recent work from \cite{ICLR2025_elrea} takes a similar approach, building ensembles with LoRA adaptors on gradient-clustered data. However, such an approach requires computing and storing gradients for every input sample, which is infeasible; consequently, they down-project the gradients into a lower dimension, losing performance. This is a general issue for calculating sample influence, such as in LESS \cite{less}.

To address this, we develop a novel online learning methodology which handles LoRA gradients in full dimensionality, thus making full use of the gradient signals. This is performed by first training a LoRA adapter \cite{hu2022lora} on a warm-up dataset to obtain meaningful gradient representations and adapt the model to the input data distribution. Next, online singular value decomposition (SVD)-based clustering is performed, where SVD over the gradient matrix is employed to estimate the number of latent clusters and initialize their centroids based on the dominant gradient directions. This ensures that clustering begins in a subspace capturing the most significant gradient variance. Centroids are then refined using an exponential moving average (EMA) update. Next, each discovered cluster is finetuned into a specialized LoRA expert.

While identifying clusters and training a LoRA expert for each is essential, a fast inference method is also needed. Techniques such as \cite{ICLR2025_elrea} employ gradient similarity during inference to identify the right set of experts; this can be very expensive, unrealistic for production deployment, and result in suboptimal accuracy. Our solution to this comes from two novel contributions: (a) we show that a correctly chosen cluster with a single expert adaptor is in fact better than an ensemble (4.2\% on average better than \citet{ICLR2025_elrea}), and (b) we introduce a fast, lightweight router trained on the resulting cluster assignments to dynamically select the most appropriate expert during inference, reducing computational cost compared to \cite{ICLR2025_elrea}, as shown in Table ~\ref{tab:comp_cost}. The three-stage pipeline of the proposed \textsc{GradientSpace} method is illustrated in Figure \ref{fig:overall}.

We evaluate \textsc{GradientSpace} across multiple domains including mathematical reasoning, code generation, finance, and creative writing across several benchmarks. Our results, as shown in Figure \ref{fig:modelacc}, produces coherent expert specialization and improved task accuracy, outperforming both state-of-the-art clustering baselines and other finetuning methods. 


\section{Background}
\label{background}



In this section, we explain preliminaries and notations we use in the rest of our paper.

\textbf{Notation.}
In this subsection we describe the notation used in this paper. For  for our setup consider the next token prediction problem. That is let $x_i \sim X$ represent a sequence of tokens, and $y_i \sim Y | X$ be the corresponding next token for the sequence $x_i$. Then the goal is to learn a function $g: x_i \mapsto y_i$ using a neural net $f_{\theta}$  parametrized by $\theta$. This is done by optimizing the loss function $\mathcal{L}$ over the dataset $\mathcal{D} = \{(x_1, y_1), (x_2, y_2), \cdots, (x_N, y_N)\}$.

\textbf{$\rho$-Lipschitz Gradient.} The loss function $\mathcal{L}$ is said to be \textbf{$\rho$-smooth} (i.e., its gradient $\nabla \mathcal{L}$ is $\rho$-Lipschitz) on the domain $\text{Range}(\theta)$ if there exists a constant $\rho > 0$ such that for all $\theta_1, \theta_2 \in \text{Range}(\theta)$:
\begin{align}
\lVert \nabla \mathcal{L}(\theta_1) - \nabla \mathcal{L}(\theta_2) \rVert \leq \rho \lVert \theta_1 - \theta_2 \rVert
\label{as:grad_lip}
\end{align}

\textbf{Bounded Gradient Variance.} Often for SGD convergence assumption under non convex settings \cite{ghadimi2013stochastic} assume the expected squared $\ell_2$-norm of the stochastic gradients is uniformly bounded by a constant $\sigma^2$. 

\textbf{Stochastic Gradient Descent (SGD).} We briefly recall the SGD update and a nonconvex convergence result following \citet{ghadimi2013stochastic}. Let $f(\theta)$ denote the (possibly nonconvex) objective and let $g_t$ be a stochastic gradient satisfying $\mathbb{E}[g_t \mid \theta_t] = \nabla \mathcal{L}(\theta_t)$ and $\mathbb{E}\!\left[\|g_t-\nabla \mathcal{L}(\theta_t)\|^2 \mid \theta_t\right] \le \sigma^2$. The SGD update is
\begin{align}
  \theta_{t+1} \;=\; \theta_t \;-\; \eta_t \, g_t .
\end{align}
Under $\rho$–Lipschitz gradients and a decreasing stepsize, e.g.\ $\eta_t = \min\!\left\{\tfrac{1}{\rho}, \tfrac{\tilde D}{\sigma\sqrt{T}}\right\}$ for $t=1,\dots,T$, if we return a random iterate $\theta_R$ uniformly from $\{\theta_1,\dots,\theta_T\}$, then
\begin{align}
  \frac{1}{\rho}\,\mathbb{E}\big[\|\nabla \mathcal{L}(\theta_R)\|^2\big]
  \;\le\; \frac{\rho D_f^2}{T} \;+\; \frac{2 D_f \sigma}{\sqrt{T}},
\end{align}
where $D_f = \sqrt{\tfrac{2(\mathcal{L}(\theta_1)-\mathcal{L}^\ast)}{\rho}}$. Hence, the expected gradient norm decreases at rate $O(1/\sqrt{T})$ in the nonconvex setting.

\section{Related Work}
\label{sec:relatedwork}

In this section, we categorize prior work into four categories: 
Training Dynamics and Gradient Alignment, Gradient-based data selection, Dataset Partitioning, Mixture of LoRA experts and compare \textsc{GradientSpace} with each of them.

\textbf{Training Dynamics and Gradient Alignment.} Gradient interference occurs when different training examples induce conflicting updates, reducing learning efficiency in multi-task or heterogeneous datasets.  \cite{pcgrad} and ~\cite{cagrad}, project gradients to mitigate conflicts, while ~\cite{recon} creates task-specific branches for high-conflict layers. However, these methods require explicit task labels to identify and resolve conflicts. To that end, token-level techniques such as ~\cite{stgc} analyze intra-expert gradient conflicts without task labels but rely on specialized routing and architecture. \textsc{GradientSpace} takes a data-centric approach where we partition the dataset so each expert receives gradient-aligned examples requiring no task labels. Moreover, ~\cite{stgc} is complementary to \textsc{GradientSpace} and can be applied alongside our method for improved performance.

\textbf{Gradient-based Data Selection.} There have been various data selection methods proposed to identify high quality subsets from large heterogeneous datasets. ~\cite{clusterucb} employs gradient-based clustering to select training samples, using cosine similarities of gradients computed with respect to pretrained models to identify influential data points. Similarly, ~\cite{tagcos} introduces task-agnostic gradient clustered coreset selection, applying K-means clustering on gradient features to gather samples with similar characteristics for efficient instruction tuning. Other gradient-based selection approaches~\cite{less, datainf, nagaraj2025} include influence function methods~\cite{infl} that prioritize samples based on Hessians and loss trajectories. While these methods demonstrate usage of gradient information for identifying important training samples, they fundamentally differ from our approach in that they select subsets of data rather than partition the entire dataset for training.

\textbf{Gradient-based clustering.} ELREA~\cite{ICLR2025_elrea} proposed clustering data based on gradients to train expert adapters and combine them with weighted averaging. During inference, they compute the gradient of the input and route to an ensemble of experts. However, there are two issues. First, ELREA relies on random projections to lower gradient dimension which causes information loss. Second, their approach requires computing gradients during inference which is computationally expensive for each input and multiple inference passes since they utilize expert ensembles. In contrast, \textsc{GradientSpace} clusters samples based on gradients using an online SVD-based clustering method utilizing full gradient dimension. This ensures that each expert learns from gradient aligned examples, allowing us to route to a single optimal expert during inference. We also design a lightweight router to determine the most optimal expert to route to during inference, leading to negligible latency overhead.

\textbf{Dataset Partitioning.} Several works have explored partitioning instruction datasets into coherent groups for improved training efficiency. ~\cite{commonit} introduces commonality-aware instruction tuning, partitioning datasets using three distinct metrics: task clustering based on task labels, embedding clustering using semantic embeddings, and length clustering based on response characteristics. Beyond topical or length cues, ~\cite{car} first filters high-quality pairs, then applies embedding-based clustering to form diverse partitions for tuning. These semantic-based partitioning approaches focus on surface-level similarities between instructions, such as topic coherence or syntactic patterns. However, they do not consider the optimization dynamics and interference that arise during gradient-based training.

\textbf{Mixture of LoRA experts.} Mixture of LoRA Experts~\cite{wu2024mixture} and its variants ~\cite{mode}, ~\cite{more}, ~\cite{mixlora}, and ~\cite{hdmole} extends parameter efficient finetuning by combining multiple LoRA adapters through learned gating and routing to enable task specialization within a single model. These methods include features such as hierarchical gating, adaptive rank selection, and load balanced expert routing to merge or activate adapters efficiently. In contrast, \textsc{GradientSpace} adopts a data-centric approach; instead of defining how experts are combined, we determine what data should each expert be trained on by clustering samples in gradient space. This ensures that each expert learns from gradient-aligned examples without relying on predefined task boundaries.

\begin{figure*}[!t]
  \centering
  \includegraphics[width=\textwidth]{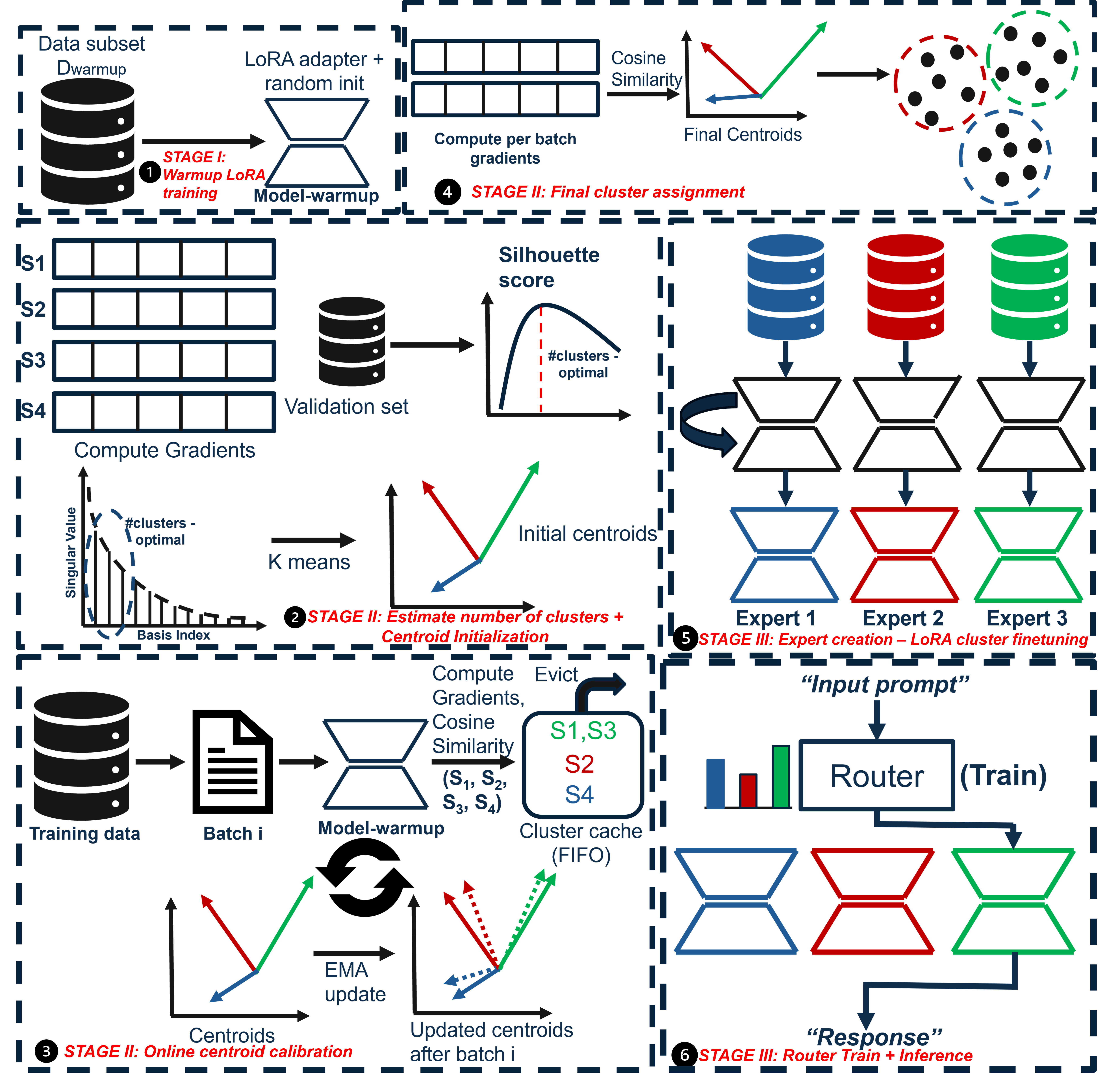}
  \caption{\textsc{GradientSpace} framework. Stage I (LoRA warm-up): train a small LoRA adapter on a warm-up split to obtain meaningful, low-dimensional gradient representations. Stage II (online SVD clustering): estimate K via SVD on a validation gradient matrix, initialize centroids in the dominant gradient subspace, then refine them online with a cluster cache and EMA updates to form gradient-aligned partitions. Stage III (experts + router): finetune one LoRA expert per cluster and use a lightweight encoder-based router to select the most appropriate expert at inference. }
  \label{fig:overall}
\end{figure*}

\section{Theoretical Formulation}
\label{sec:theory}
In this section, we present a formal analysis supporting the claims of the \textsc{GradientSpace} framework. Before we present our results we define a few terms. Let $\mathcal{D}$ be a dataset, and let $\{\mathcal{D}_1, \mathcal{D}_2, \ldots, \mathcal{D}_k\}$ be a partition of $\mathcal{D}$ (i.e., $\bigcup_{i=1}^k \mathcal{D}_i = \mathcal{D}$ and $\mathcal{D}_i \cap \mathcal{D}_j = \emptyset$ for $i \neq j$). Let $\{\nabla \mathcal{L}_i(\theta)\}_{i=1}^k$ be the gradient vectors for these subsets.

\begin{Lemma} \label{lem:var-decomp-simple}
Let $\alpha_1,\dots,\alpha_k \in [0, 1]$ such that $\sum_{i=1}^k \alpha_i = 1$. Define the combined gradient $\nabla \mathcal{L}(\theta)$ as:
\[
\nabla \mathcal{L}(\theta) := \sum_{i=1}^k \alpha_i \nabla \mathcal{L}_i(\theta).
\]
Then the variance of the combined gradient is given by:
\begin{equation}
\begin{split}
\mathrm{Var}(\nabla \mathcal{L}(\theta))
= {} & \sum_{i=1}^k
\alpha_i^2 \,
\mathrm{Var}(\nabla \mathcal{L}_i(\theta))
\\
& \;+\;
\sum_{i \neq j}
\alpha_i \alpha_j \,
\mathrm{Cov}\big(
\nabla \mathcal{L}_i(\theta),
\nabla \mathcal{L}_j(\theta)
\big)
\end{split}
\end{equation}
\end{Lemma}

\textbf{Sketch of Proof.}
At a high level, the result follows from the standard variance decomposition rule. The combined gradient is a weighted sum of the gradients from each subset, so its total variance naturally splits into two parts: the weighted sum of the individual (within-subset) variances and the cross-term capturing how gradients from different subsets co-vary. This yields the stated expression for the variance of the combined gradient. The proof is provided in the Appendix.

\textbf{Interpreting Theory.}
The lemma shows that training on a mixture of subsets aggregates two noise sources: the intrinsic gradient variance within each subset and the cross-subset covariance. In the \textsc{GradientSpace} setting, routing to coherent subsets reduces within-subset dispersion and suppresses harmful cross-covariances, recovering the “experts converge faster” intuition: per-subset (expert) updates omit the cross terms, yielding lower gradient variance and tighter expected $\varepsilon$-stationarity guarantees than a shared model that mixes conflicting gradients.

Next, we establish a result showing that training on subsets identified by \textsc{GradientSpace} yields a reduction in gradient variance.
\begin{Theorem}[Variance Reduction] \label{th:var_red}
Let $\Var(d)$ denote the variance of the gradients computed on a dataset $d$. Let a dataset $\mathcal{D}$ be partitioned into subsets $\{\mathcal{D}_1, \mathcal{D}_2, \ldots, \mathcal{D}_k\}$ using the \textsc{GradientSpace} method. Then for any subset $\mathcal{D}_i$ in the partition:
\begin{align}
    \Var\left(\mathcal{D}_i \right) \leq \Var(\mathcal{D}) \quad \text{for all } i \in \{1,\ldots,k\}.
\end{align}
\end{Theorem}

\textbf{Sketch of Proof.}  
The result follows from the law of total variance. When the dataset $\mathcal{D}$ is partitioned into subsets $\{\mathcal{D}_i\}$ using the \textsc{GradientSpace} method, the total gradient variance decomposes into within-subset and between-subset components. The algorithm explicitly minimizes within-subset dispersion in gradient space, ensuring that each $\mathcal{D}_i$ has gradients that are more homogeneous than the global set. Consequently, the within-subset variance $\Var(\mathcal{D}_i)$ cannot exceed the overall variance $\Var(\mathcal{D})$, and is strictly smaller whenever subsets differ in their mean gradient directions. The proof is provided in the Appendix.

\textbf{Interpreting Theory.}  
This result shows that partitioning data in gradient space effectively separates examples with conflicting gradient directions, thereby reducing gradient noise during optimization. Lower gradient variance leads to more stable updates and faster convergence under stochastic gradient descent. In this way, training on subsets identified by \textsc{GradientSpace} mirrors the advantage of expert models over shared models, each subset behaves as a low-variance expert, accelerating overall training dynamics.

Finally, we present the analysis on improved $\varepsilon$-stationarity when using \textsc{GradientSpace}.

\textbf{Definition (Expected $\varepsilon$-Stationarity).}  
An SGD training run is said to be \emph{expected $\varepsilon$-stationary} after $T$ steps if
\[
\frac{1}{T} \sum_{t=0}^{T-1}
\mathbb{E}\big[\|\nabla \mathcal{L}(\theta_t)\|^2\big] \leq \varepsilon,
\]
that is, the average expected squared gradient norm across its iterates is at most $\varepsilon$. This indicates that, on average, the run has reached an $\varepsilon$-approximate stationary point of the loss surface. This definition is a modification of the standard $\varepsilon$-stationarity notion, where only the final iterate satisfies $\|\nabla \mathcal{L}(\theta_T)\|^2 \leq \varepsilon$. The updated definition simplifies formal analysis of non-convex SGD training.

We define the variance of the stochastic gradients for the full dataset and for each cluster $i$ as:
\[
\sigma_{\mathcal{D}}^2 = \Var(\nabla \mathcal{L}_{\mathcal{D}}(\theta)),
\qquad
\sigma_{\mathcal{D}_i}^2 = \Var(\nabla \mathcal{L}_{\mathcal{D}_i}(\theta)).
\]
From Theorem \ref{th:var_red}
 we have $\bar{\sigma}_{\text{cluster}}^2 \le \sigma_{\mathcal{D}}^2$. Let $\mathsf{SGD}_{\mathcal{D}}$ and $\mathsf{SGD}_{\text{cluster}}$ be defined as above. Assume both algorithms
    (i) start from the same initialization $\theta_0$,
    (ii) use the same stepsize $\eta \le 1/\rho$,
    (iii) and are run for the same number of iterations $T$ on an $\rho$-smooth objective $\mathcal{L}$ with unbiased gradient estimates. We group these three under the \textit{SGD convergence assumptions}.
Let $\mathcal{L}^\star := \inf_\theta \mathcal{L}(\theta)$.
Then their expected $\varepsilon$-stationarity levels after $T$ steps satisfy
\begin{align}
\varepsilon_{\mathcal{D}}
&\;\le\;
\frac{\eta \rho}{2T}\big(\mathcal{L}(\theta_0) - \mathcal{L}^\star\big)
\;+\;
\rho \eta \sigma_{\mathcal{D}}^2, \\
\varepsilon_{\text{cluster}}
&\;\le\;
\frac{\eta \rho}{2T}\big(\mathcal{L}(\theta_0) - \mathcal{L}^\star\big)
\;+\;
\rho \eta \bar{\sigma}_{\text{cluster}}^2.
\end{align}

As $T \to \infty$, the initial suboptimality gap term vanishes, making the asymptotic convergence behavior dominated by the variance terms. 
The ratio of these asymptotic bounds is
\[
\lim_{T \to \infty} \frac{\varepsilon_{\text{cluster}}}{\varepsilon_{\mathcal{D}}}
\;\le\;
\frac{\rho \eta \bar{\sigma}_{\text{cluster}}^2}{\rho \eta \sigma_{\mathcal{D}}^2}
=
\frac{\bar{\sigma}_{\text{cluster}}^2}{\sigma_{\mathcal{D}}^2}.
\]
Since $\bar{\sigma}_{\text{cluster}}^2 \le \sigma_{\mathcal{D}}^2$, this ratio is less than or equal to 1, demonstrating the asymptotic improvement in convergence to a stationary point. Hence we can write the follwing result.

\begin{Theorem}[Asymptotic Stationarity Improvement]
\label{th:asymptotic-stationarity}
Under the SGD convergence assumptions, the ratio of the expected $\varepsilon$-stationarity bounds for $\mathsf{SGD}_{\text{cluster}}$ and $\mathsf{SGD}_{\mathcal{D}}$ satisfies
\[
\lim_{T \to \infty} \frac{\varepsilon_{\text{cluster}}}{\varepsilon_{\mathcal{D}}}
\;\le\;
\frac{\rho \eta \bar{\sigma}_{\text{cluster}}^2}{\rho  \eta \sigma_{\mathcal{D}}^2}
=
\frac{\bar{\sigma}_{\text{cluster}}^2}{\sigma_{\mathcal{D}}^2}.
\]
\end{Theorem}


\textbf{Interpreting Theory.}
Theorem~\ref{th:asymptotic-stationarity} shows that in the asymptotic case, SGD driven by \textsc{GradientSpace} gradients is \emph{provably closer to first-order stationarity} (i.e., has a smaller bound) than vanilla SGD on the full dataset. Equivalently, \textsc{GradientSpace} reaches an $\varepsilon$-stationary solution that is: it guarantees a lower expected gradient norm at the same training cost.

\section{Methodology}
\label{sec:method}

We introduce \textsc{GradientSpace}, a framework that partitions heterogeneous instruction tuning data based on \textit{gradient similarity} without any predefined task labels. \textsc{GradientSpace} operates in three stages. In \textbf{Stage I}, we train a LoRA adapter~\cite{hu2022lora} using a small subset of data to adapt the model to the input data distribution. In \textbf{Stage II}, we group input samples with similar gradient updates using an online SVD-based clustering algorithm. Finally, in \textbf{Stage III}, we finetune the LoRA adapter from stage I for each discovered cluster to create experts and train a lightweight router to dynamically select the appropriate expert during inference. Figure~\ref{fig:overall} provides an overview of the proposed framework. We first explain the problem setup and describe each stage in detail below.

\subsection{Problem Setup}

Let $\mathcal{D} = \{(x_i, y_i)\}_{i=1}^N$ be a finetuning dataset consisting of heterogeneous instruction samples. 
Given a model $f_\theta$ with parameters $\theta$, the gradient of the loss $\mathcal{L}$ for a sample $(x_i, y_i)$ is
\[
g_i = \nabla_\theta \mathcal{L}(f_\theta(x_i), y_i).
\]
Two samples $i, j$ are considered \textit{gradient-aligned} if their cosine similarity
\[
\text{sim}(g_i, g_j) = \frac{g_i^\top g_j}{\|g_i\|\|g_j\|}
\]
is greater than a threshold. 
Intuitively, these samples push the model parameters in similar directions and can be optimized together without destructive interference. 
The objective of \textsc{GradientSpace} is to partition $\mathcal{D}$ into $K$ clusters 
$\{\mathcal{D}_1, \mathcal{D}_2, \dots, \mathcal{D}_K\}$ that maximize intra-cluster gradient alignment while minimizing cross-cluster conflict. Formally, we define the objective as follows:

\begin{equation}
\begin{aligned}
    & \underset{\{D_k\}_{k=1}^{K}}{\text{maximize}}
    & & \sum_{k=1}^{K} \sum_{i,j \in D_k} \text{sim}(g_i, g_j) \\
    & \text{subject to}
    & & \sum_{p \neq q} \sum_{i \in D_p,\, j \in D_q} \text{sim}(g_i, g_j) \text{ minimized.}
\end{aligned}
\end{equation}

\subsection{Stage I: Gradient Approximation via LoRA Warm-up}





We adopt the standard low rank adaptation (LoRA) method \citep{hu2022lora} to reduce the number of trainable parameters and enable more efficient gradient computations. LoRA achieves this by freezing the pretrained model weights and injecting trainable low-rank adapters into the linear layers distributed throughout the network. We use LoRA to instruction tune a pre-trained model (e.g., LLaMA-2-7B) on a small random subset $\mathcal{D}_{\rm warmup} \subset \mathcal{D}$ (5\% of the training dataset) for a few epochs, training only the low-rank adapters $\Delta \theta_{\rm LoRA}$. For each example $(x_i,y_i) \in \mathcal{D}_{\rm warmup}$, we compute the gradient with respect to the LoRA parameters:
\[
\tilde g_i = \nabla_{\Delta \theta_{\rm LoRA}} L\big(f_{\theta,\Delta\theta_{\rm LoRA}}(x_i),y_i\big),
\]
which serves as a compact, low-dimensional representation of the full-parameter gradient.We primarily perform this step to obtain meaningful gradients in the next stages based on the input data distribution.

\subsection{Stage II: Online SVD-based Clustering and Centroid Refinement}
In this stage, we estimate the number of clusters for \textsc{GradientSpace}, initialize centroids using SVD and perform online gradient clustering to determine per sample alignment and refine the initialized centroids.
\subsubsection{Estimating number of clusters and centroid initialization}

After the LoRA warm-up stage (Stage I), we now compute the gradient representations $\{\tilde g_i\}_{i=1}^{n_{\rm val}}$ for a small validation subset. These gradients capture the base model updates and serve as the input for identifying latent task structure. To organize the gradients into coherent groups, we represent each cluster by a \textit{centroid} $c_k \in \mathbb{R}^d$, defined as the mean of all gradients assigned to that cluster. Each centroid thus represents the “average update direction” for a latent task, providing a compact summary of the gradients in that group and serving as a reference for clustering new examples.

To estimate the optimal number of clusters $K$, we perform singular value decomposition (SVD) on the gradient matrix $G \in \mathbb{R}^{n_{\rm val} \times d}$, where each row corresponds to a gradient:
\[
G = U \Sigma V^\top,
\]
with $\Sigma = \mathrm{diag}(\sigma_1, \sigma_2, \dots, \sigma_r)$ containing the singular values. The explained variance ratio for the top-$k$ components is
\[
\mathrm{ExplainedVariance}(k) = \frac{\sum_{i=1}^k \sigma_i^2}{\sum_{i=1}^r \sigma_i^2},
\]
quantifying the fraction of total gradient variance captured by the dominant directions. 

For a range of variance thresholds (e.g., 80\%--95\%), we project the gradients onto the corresponding top-$k$ singular vectors:
\[
\tilde G_k = G V_k,
\]
where $V_k \in \mathbb{R}^{d \times k}$ contains the top-$k$ right singular vectors. We then apply K-means clustering in this subspace, and the \textbf{Silhouette score} is computed to evaluate cluster coherence. The number of clusters $K$ that yields the highest Silhouette score is selected as the optimal cluster count. We note that the highest silhoutte score corresponds to the best accuracy as shown in section~\ref{sec:expt}. Finally, the centroids $\{c_k\}_{k=1}^K$ obtained from K-means in the SVD-projected space serve as the initial “prototype gradients” for each latent task, providing the starting points for subsequent online clustering across the full training dataset.

\subsubsection{Online Centroid Caliberation}

After computing the initial centroids in the previous step, we perform online clustering to efficiently manage gradient assignments across the full dataset without storing all gradients simultaneously. For each incoming mini-batch of examples $\mathcal{B}_t = \{(x_i, y_i)\}$, we compute the corresponding set of low-dimensional gradients $\tilde{G}_t = \{\tilde{g}_i\}_{i \in \mathcal{B}_t}$ and assign each gradient to the nearest centroid based on cosine similarity.

To maintain stable and adaptive clusters, we introduce a \emph{cluster cache} that records recent gradient assignments for each centroid.  This ensures the order of samples do not influence centroid creation disproportionately, which can occur if updates are computed solely from the current batch. By maintaining a rolling window of the most recent gradients assigned to each cluster, the centroids are re-evaluated using the average of the cached gradients. We implement the cluster cache as a fixed-size buffer for each cluster, discarding older entries in a first-in-first-out manner to ensure the centroids reflect recent and important gradient information. 

The centroids themselves are updated incrementally using an exponential moving average (EMA). Formally, let $C_k$ denote the fixed-size buffer (cluster cache) maintaining the most recent gradients assigned to cluster $k$,:
\[
c_k^{(t+1)} = \beta c_k^{(t)} + (1 - \beta) \frac{1}{|\mathcal{C}_k|} \sum_{g \in \mathcal{C}_k} \tilde{g},
\]
where $\beta \in [0,1)$ controls the temporal smoothing. The cluster cache size for each cluster $k$ is set to $|\mathcal{C}_k| = \alpha \cdot |\mathcal{B}_t|$, 
where $|\mathcal{B}_t|$ denotes the batch size and $\alpha \in \mathbb{N}$ is a tunable scaling parameter.
We find that values of $\alpha$ in the range $[5, 10]$ yield robust performance across diverse tasks, balancing stability and memory overhead. This refinement process is repeated for several epochs until the centroids converge, yielding the final set of stable centroids $\{ c_k^{(\text{final})} \}_{k=1}^K$.

\subsubsection{Final Cluster Assignment}
After centroid refinement converges, we perform a final assignment pass over all samples to establish fixed cluster assignments. For each training example $(x_i, y_i)$ with gradient representation $\tilde{g}_i$, we compute cosine similarity with all centroids and assign it to the most aligned one:
\begin{equation}
k_i = \arg\max_{k \in \{1, \dots, K\}} \; \text{sim}(\tilde{g}_i, c_k^{(\text{final})}),
\end{equation}
where $\text{sim}(\cdot,\cdot)$ denotes cosine similarity. 
This produces the final partition $\{D_1, D_2, \dots, D_K\}$, where each subset contains examples that induce mutually reinforcing gradient updates. The resulting assignments are subsequently used in Stage~III to train cluster-specific LoRA experts and to supervise the router.

\subsection{Stage III: Expert Specialization and Adaptive Routing}
In this stage, we create experts by finetuning LoRA adapters (from Stage I) and train a router to select the optimal expert during inference.
\subsubsection{Cluster-Specific Expert Fine-Tuning}
After online clustering, each cluster $\mathcal{D}_k$ forms the data foundation for a specialized LoRA expert. 
We instantiate $K$ LoRA adapters $\{\Delta \theta_{\text{LoRA}}^{(k)}\}_{k=1}^K$ and finetune each expert on its corresponding subset
\[
\min_{\Delta \theta_{\text{LoRA}}^{(k)}} \frac{1}{|\mathcal{D}_k|} \sum_{(x_i, y_i) \in \mathcal{D}_k} \mathcal{L}(f_{\theta, \Delta \theta_{\text{LoRA}}^{(k)}}(x_i), y_i).
\]
Each expert captures a distinct “skill” in the model’s parameter space, reflecting the structure of the discovered gradient clusters. 
Importantly, the base model remains frozen across all experts, thereby ensuring low training overhead. We note that we use the warmup up lora adapter rather than randomly initialized adapter since this helps in improving performance. 

\subsubsection{Router Training and Inference}

\begin{figure}[htb]
  \includegraphics[width=\columnwidth]{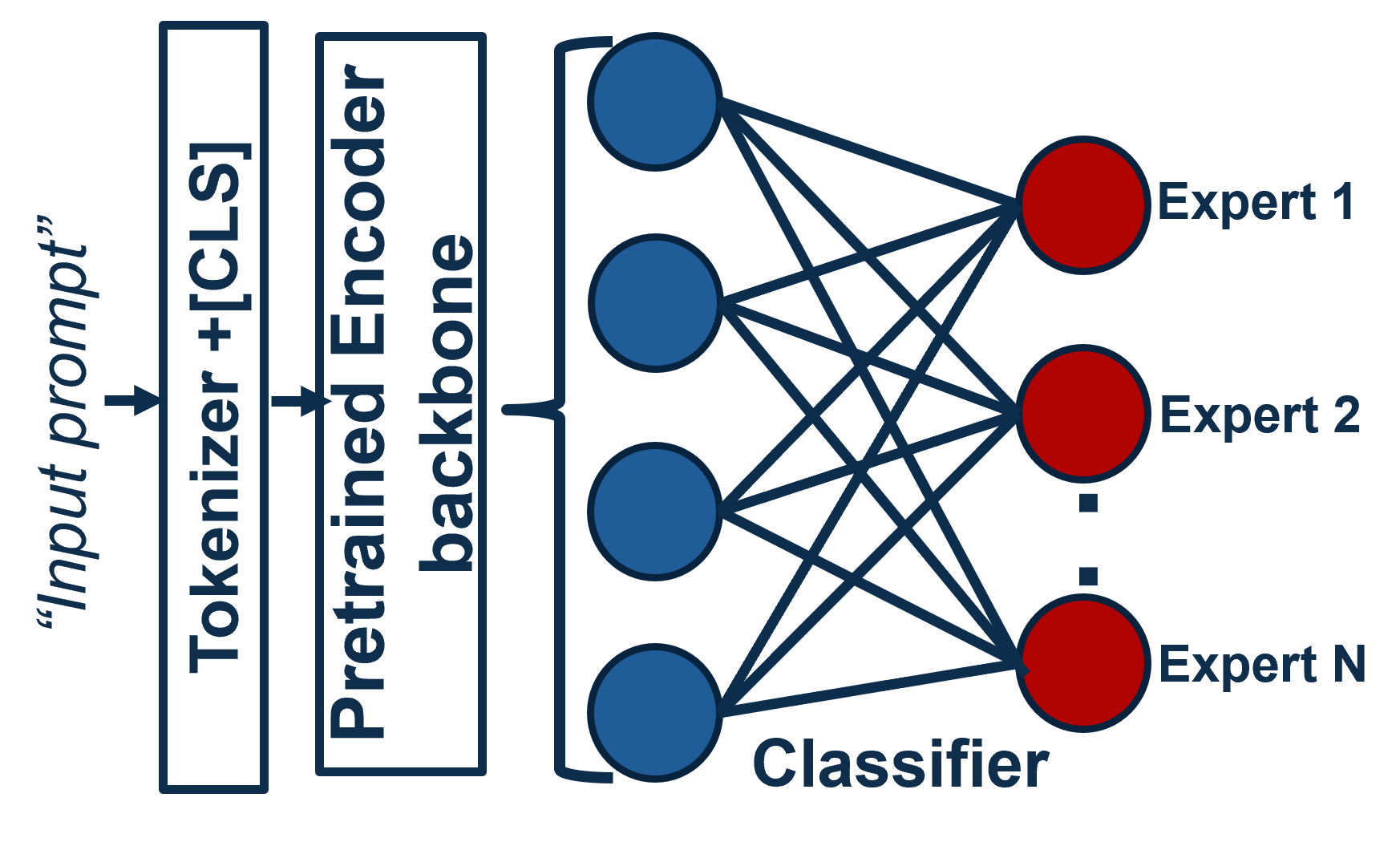}
  \caption{\textsc{GradientSpace} Router. }
  \label{fig:router}
\end{figure}

We design a lightweight router that predicts the most appropriate expert for each input. As illustrated in Figure~\ref{fig:router}, the router is implemented as an encoder-based model in which the standard classification head is replaced with a linear projection layer of dimension $K$, corresponding to the number of experts. During training, the encoder backbone is frozen, and only the final projection layer is updated. 

Let $h_i$ denote the fixed input representation extracted from the encoder (e.g., the final-layer hidden state or gradient-derived feature) corresponding to input $x_i$. The router takes $h_i$ as input and outputs a distribution over clusters:
\[
p_\phi(k \mid x_i) = \text{softmax}(W h_i + b),
\]
where $\phi = \{W, b\}$ are router parameters. 
It is trained using cross-entropy loss with the discovered cluster assignments as supervision:
\[
\mathcal{L}_{\text{router}} = - \sum_i \log p_\phi(k_i \mid x_i).
\]
During inference, the router assigns each input to the most relevant expert, and the final output is produced by that expert’s LoRA adapter:
\[
\hat{y} = f_{\theta, \Delta \theta_{\text{LoRA}}^{(k^\ast)}}(x), \quad k^\ast = \arg\max_k p_\phi(k \mid x).
\]
We present the accuracy and latency results of this router compared to other state-of-the-art methods in section~\ref{subsec:routexp}.

\section{Experiments}
\label{sec:expt}
In this section we details the experiments to validate the claims of the paper.

\subsection{Gradient vs Input Space}
One of the motivations for this paper is that input similarity is not the same as learning similarity. In this section, we provide evidence for that claim. We test this hypothesis using the Llama 2-7B model and the mteb/stsbenchmark-sts dataset~\cite{mteb}. This dataset consists of pairs of inputs with varying degrees of semantic similarity, where similarity between each pair is measured using cosine similarity between their embeddings. We show that high cosine similarity between a pair of inputs does not necessarily correspond to similar learning behavior.

\textbf{Experiment}
We evaluate 500 human-annotated pairs and measure the gradients of the LoRA weights for Llama 2-7B and compute (i) cosine similarity between their input embeddings and (ii) cosine similarity between the corresponding LoRA gradients obtained from the model’s parameter updates. 

\begin{figure}[htb]
  \includegraphics[width=\columnwidth]{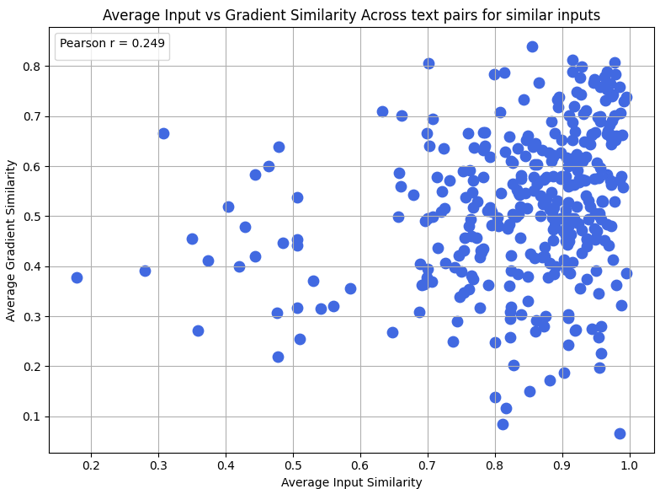}
  \caption{Comparison of input embedding similarity and gradient similarity across 500 randomly selected instruction pairs from the MTEB STS benchmark dataset using LLaMA-2-7B.}
  \label{fig:motiv}
\end{figure}

\textbf{Results and Takeaways}
As shown in Figure ~\ref{fig:motiv}, our results indicate that gradient similarity, which reflects how the model would update its parameters, is not predicted by input embedding similarity (shows poor correlation) confirming our previous hypothesis.\\

\begin{table*}[t]
\centering
\small
\setlength{\tabcolsep}{4pt}
\begin{tabular}{lcc ccc ccc}
\toprule
\multirow{2}{*}{\textbf{Method}} &
\multicolumn{2}{c}{\textbf{Data Mix}} &
\multicolumn{2}{c}{\textbf{GSM8K}} &
\multicolumn{2}{c}{\textbf{MATH}} \\
\cmidrule(lr){2-3} \cmidrule(lr){4-5} \cmidrule(lr){6-7}
& \textbf{LLaMA-3.2} & \textbf{LLaMA-2} &
  \textbf{LLaMA-3.2} & \textbf{LLaMA-2} &
  \textbf{LLaMA-3.2} & \textbf{LLaMA-2} \\
\midrule
Zero-shot prompting (Pretrained)                  & 20.1 & 33.4 & 31.3 & 40.6 & 12.1 & 18.8\\
Finetuning              & 33.8 & 48.9 & 41.5 & 50.8 & 23.1 & 35.4 \\
LoRA Finetuning        & 33.6 & 48.1 & 40.7 & 47.6 & 20.4 & 28.4 \\
Random clustering + LoRA             & 26.9 & 38.3 & 31.2 & 35.4 & 15.6 & 24.3 \\
K-means + LoRA (Mixture of LoRA)    & 38.8 & 53.3 & 44.8 & 53.1 & 19.3 & 32.9 \\
Model Merging                        & 34.1 & 47.1 & 30.6 & 40.7 & 15.9 & 27.8 \\
ELREA ~\cite{ICLR2025_elrea}   & 41.7 & 56.2 & 43.2 & 52.4 & 22.2 & 41.8 \\

\rowcolor{gray!10}
\textsc{GradientSpace} + SVD init (Ours)      & \textbf{46.4} & \textbf{59.1} & \textbf{46.7} & \textbf{54.7} & \textbf{28.9} & \textbf{46.8} \\
\bottomrule
\end{tabular}
\caption{
Accuracy (\%) of different finetuning methods on three benchmarks using \textbf{LLaMA-3.2-1B} and \textbf{LLaMA-2-7B} models. 
The best result for each dataset is shown in \textbf{bold}, and the second best is \underline{underlined}.
}
\label{tab:accuracy_comparison}
\end{table*}

\subsection{Instruction Tuning Experiments}
\label{sec:instruction_tuning}



We conduct experiments using two pretrained base models: \textsc{LLaMA-2-7B}~\cite{llama2} and \textsc{LLaMA-3.2-1B}~\cite{llama3}. 
Both models are instruction-tuned under identical training configurations for fair comparison. For LoRA, we use \verb|r=32| for all our experiments. We evaluate performance across three datasets: a mixed-domain corpora (Data Mix), \textsc{GSM8K}~\cite{gsm8k} and the \textsc{MATH} benchmark~\cite{nlile_math_benchmark}. 

\noindent\textbf{Data Mix.}
This consists of four distinct domains---\textit{finance}, \textit{creative writing}, \textit{code generation}, and \textit{mathematical reasoning}---sampled respectively from 
\verb|gbharti/finance-alpaca|~\cite{gbharti_finance_alpaca}, 
\verb|roneneldan/TinyStories|~\cite{roneneldan_tinystories}, 
\verb|iamtarun/code_instructions_120k_alpaca|
\newline~\cite{iamtarun_codeinstructions}, 
\verb|nlile/hendrycks-MATH-benchmark|
\newline~\cite{nlile_math_benchmark}. 
We sample 3{,}000 examples from each domain to maintain balance and diversity across instruction styles.

Table~\ref{tab:accuracy_comparison}. For a fair comparison, we consider the following baselines:

\begin{itemize}
    \item \textbf{Zero shot prompting.} The unmodified pretraining, without any finetuning.
    \item \textbf{Finetuning.} A single model obtained by finetuning the model on the entire dataset.
    \item \textbf{LoRA Finetuning.} The same training setup as the previous baseline, but instead of updating all model parameters, we only train LoRA adapters.
    \item \textbf{Random clustering + LoRA.} We randomly partition the corpus into a few clusters. For each random cluster we train a distinct LoRA adapter. At inference time we use our router to select which adapter to apply to a given query.
    \item \textbf{K-means + LoRA (Mixture of LoRA).} We embed each training sample, cluster the samples into 5 clusters using K-means over these input embeddings, train one LoRA adapter per cluster, and again use the router at inference time. We perform this experiment three times and report average accuracy.
    \item \textbf{Model merging.} Independently finetuned LoRA adapters (each trained on a disjoint subset of the data) are merged into a single model by linearly combining their weights.
    \item \textbf{ELREA.~\cite{ICLR2025_elrea}} This method partitions the dataset by clustering gradient representations that are compressed to d=8192 using random projection. At inference time, we follow their optimal setting by routing inputs to the full ensemble of experts, aggregating their predictions via a weighted average derived from softmax-normalized gradient alignment scores.
    \item \textbf{\textsc{GradientSpace} (Ours).} Our method, in which clusters are defined using gradient similarity. We form clusters by grouping samples whose gradients point in similar update directions, and then train one LoRA adapter per gradient-aligned cluster. Each cluster is partitioned based on our online SVD-based clustering algorithm utilizing full gradient dimension. At inference time, we have a tiny encoder-based classifier to chose the most appropriate expert.

\end{itemize}


\textbf{Results.} Table~\ref{tab:accuracy_comparison} summarizes results across three benchmarks—Data Mix, GSM8K, and MATH using both LLaMA-3.2-1B and LLaMA-2-7B models. We observe that creating multiple experts with random clustering is ineffective since the accuracy is only 38.3 in DataMix and 24.3 in MATH for example. Next, clustering by input embeddings offers a stronger baseline (i.e.) K-means + LoRA reaches 53.3 on Data Mix and 32.9 on MATH which suggests that routing to specialized adapters helps, though input similarity alone fails to capture how the model actually learns. Furthermore, we explicitly compare our method to ELREA~\cite{ICLR2025_elrea}, which employs a weighted ensemble of experts. We find that \textsc{GradientSpace} outperforms ELREA by an average of 4.2\% across all benchmarks, with the most significant gains observed in the MATH domain where our method achieves 22.2\% compared to ELREA’s 28.9\% (on LLaMA-2). This indicates that ELREA’s strategy of random projection to lower dimension introduces information loss, whereas our approach of strictly routing to a single, SVD-initialized expert effectively isolates conflicting gradients and maximizes specialization. In contrast, \textsc{GradientSpace} consistently achieves the highest accuracy across all tasks.

\textbf{Takeaways.}
First, we observe that grouping examples by gradient alignment produces better specialization than grouping by input embeddings, supporting the claim that input similarity is not the same as learning similarity. Next, we also observe that SVD initialization plays a crucial role in \textsc{GradientSpace} in identifying gradient conflicts; incorporating SVD-based gradient initialization provides consistent accuracy gains over ELREA~\cite{ICLR2025_elrea}. This proves that gradient-based data clustering is a promising direction for scalable instruction tuning.
\vspace{-12pt}
\begin{figure}[ht]
    \centering
    \includegraphics[width=1\linewidth]{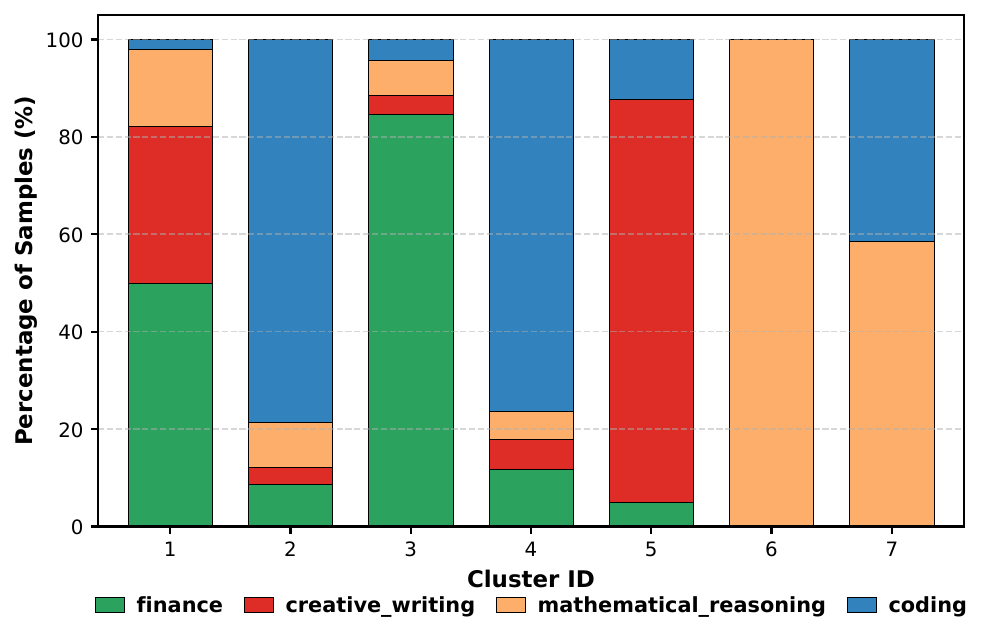}
    \vspace{-8pt}
    \caption{Cluster distribution for Data Mix.}
    \vspace{-8pt}
    \label{fig:cluster_analysis}
\end{figure}

\subsection{Clustering Analysis}
In this section we analyze the effect of applying the \textit{\textsc{GradientSpace}} framework to DataMix dataset. The resulting cluster assignments are shown in Figure~\ref{fig:cluster_analysis}.

~\textbf{Results} Figure~\ref{fig:cluster_analysis} shows that the discovered clusters are not strictly homogeneous by source domain. Even though the data are labeled into clear domains (finance, creative writing, mathematical reasoning, and coding), \textsc{GradientSpace} groups samples across these domains when their gradients are aligned, i.e., when they reinforce similar model updates. To illustrate the structure of these clusters, we compute TF–IDF statistics within each cluster and list high-weight terms in Table~\ref{tab:cluster_dist}.

\begin{table}[h]
\centering
\resizebox{\linewidth}{!}{
\begin{tabular}{l p{0.8\linewidth}}
\hline
\textbf{Cluster} & \textbf{Top Terms} \\
\hline
Cluster 1 &
explain, because, however, example, what, how \\
Cluster 2 &
function, list, return, if, price, number, calculate \\
Cluster 3 &
stock, market, tax, investment, loan, company, risk, shares, price \\
Cluster 4 &
def, return, int, string, for, list, array, len, self, import \\
Cluster 5 &
said, happy, little, girl, mom, day, once, time, Lily, Tom \\
Cluster 6 &
frac, sqrt, equation, angle, number, sum, proof, \textbackslash cdot, \textbackslash align \\
Cluster 7 &
prime, number, matrix, array, loop, if/else, calculate \\
\hline
\end{tabular}
}
\caption{High-weight TF--IDF terms per cluster.}
\label{tab:cluster_dist}
\end{table}

These terms indicate that clusters are organized around functional roles rather than strictly around dataset labels. For example, Cluster 2 mixes procedural language (“function,” “return,” “if”) with numerical/computational terms (“price,” “number,” “calculate”), which spans both coding and finance. Cluster 6 and Cluster 7 both show mathematical structure, but with different emphases: Cluster 6 is associated with formal mathematical presentation (symbols such as $\cdot$ and LaTeX-style tokens), whereas Cluster 7 emphasizes algorithmic reasoning over mathematical objects (“matrix,” “array,” “loop,” “if/else,” “calculate”), which bridges math reasoning and code.

\textbf{Takeaways.}
\textsc{GradientSpace} discovers cross-domain clusters when they yield compatible training signals rather than naive human-provided labels, indicating that gradient alignment captures learning similarity rather than dataset labels. The presence of mixed clusters (e.g., finance with code, math with programmatic reasoning) shows that some instruction types reinforce each other across sources.

\subsection{Router Performance}
\label{subsec:routexp}
We evaluate four router variants to study the trade-off between computational cost and accuracy:
\begin{itemize}
    \item \textbf{Keyword Similarity (KS):} a heuristic router that assigns an expert based on keyword overlap between the input prompt and expert-domain metadata.
    \item \textbf{Embedding Similarity (ES):} computes cosine similarity between the SentenceTransformer embedding ~\cite{sentencebert} (all-MiniLM-L6-v2) of the input and the mean embedding of each expert’s data.
    \item \textbf{Gradient Similarity (GS) :} matches the input prompt to experts using the stored average gradient vectors per cluster, selecting the expert with the maximum cosine alignment. This formulation is identical to the approach of \citet{ICLR2025_elrea}.
    \item \textbf{Semantic Predictor (SP) (Ours):} DistilBERT-like backbone+Linear projector router trained on $(x_i, k_i)$ pairs as described above.
\end{itemize}

\textbf{Results. and Takeaways.}
We summarize the performance of different routers in Table~\ref{tab:router_performance}. 
The proposed semantic predictor router achieves \textbf{99.4\% accuracy} with an average latency of \textbf{3.9 ms}, outperforming all other methods in achieving the best accuracy-efficiency tradeoff. Gradient similarity routing provides strong alignment but is prohibitively slow ($\sim$2000 ms) due to full gradient computation, while keyword-based heuristics are fast but unreliable (78.5\% accuracy).

\subsection{Computational Complexity}
We now compare the computational complexity of ELREA~\cite{ICLR2025_elrea} and our work during inference shown in Table ~\ref{tab:comp_cost}. Let us assume the number of floating point operations for one forward pass through a single LoRA adapter is $F_{\text{LoRA}}$, for the base model is $F_{\text{Base}}$ and for our semantic predictor is $F_{\text{SP}}$. ELREA relies on per input gradient computation to route among experts, requiring one forward and one backward pass through the LoRA parameters, followed by $k$ forward passes when ensembling the top-$k$ experts. We assume that approximately backward pass takes twice the flops of forward pass. This results in a total of $(1 + k) F_{\text{Base}} + (3 + k) F_{\text{LoRA}}$ FLOPs per query. In contrast, \textit{\textsc{GradientSpace}} uses gradients only offline to form clusters and train experts, and employs a lightweight semantic router that adds negligible overhead on top of a single expert forward, leading to approximately $F_{\text{Base}} + F_{\text{LoRA}}$ FLOPs per query.


\begin{table}[t]
\centering
\small
\setlength{\tabcolsep}{6pt}
\renewcommand{\arraystretch}{1.15}
\caption{Router design comparison in terms of accuracy and latency.}
\label{tab:router_performance}
\begin{tabular}{p{0.35\columnwidth}cc}
\toprule
\textbf{Router Type} & \textbf{Accuracy (\%)} & \textbf{Latency (ms)} \\
\midrule
KS                & 78.5 & 0.01 \\
ES               & 95.1 & 5.0 \\
GS ~\cite{ICLR2025_elrea}         & 96.4 & 2089.3 \\
\rowcolor{gray!10}
\textbf{SP (Ours)} & \textbf{99.4} & \textbf{3.9} \\
\bottomrule
\end{tabular}
\end{table}

\subsection{Ablation: Varying number of clusters}
\label{subsec:ablation}

\textbf{Experiment} We conduct an ablation study using the \textsc{MATH} dataset by varying the number of clusters $K$To analyze the effect of the number of discovered clusters on cluster quality and final accuracy.

\begin{figure}[ht]
    \centering
    \includegraphics[width=1\linewidth]{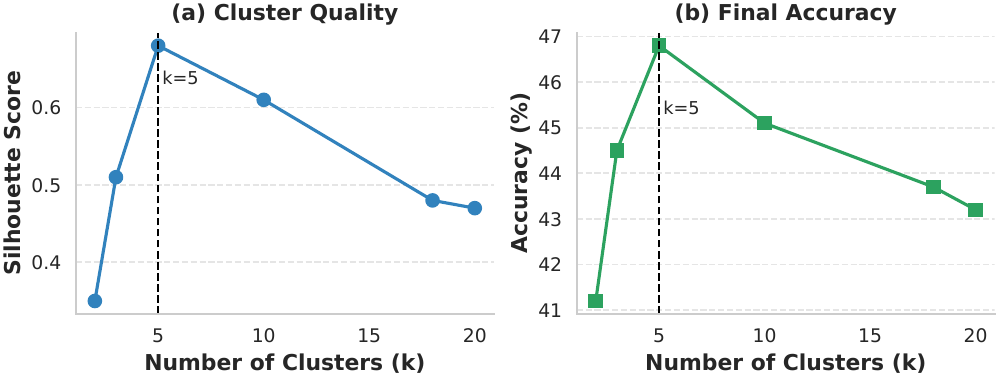}
    \vspace{-8pt}
    \caption{cluster cohesion and Downstream performance varying number of clusters}
    \vspace{-8pt}
    \label{fig:silhoutte}
\end{figure}

\textbf{Results and Takeaways} 
We show the results of this experiment in Figure~\ref{fig:silhoutte}. We observe that the cluster quality and model accuracy improve as $K$ increases up to $K{=}5$, beyond which both measures begin to plateau or slightly degrade. This indicates that excessively fine partitioning leads to redundant or noisy gradient clusters that harm generalization, 
while too few clusters underrepresent the latent structure in the data. To achieve the best trade-off between gradient alignment and downstream performance, we obtain the optimal number of clusters using the validation as described in Section ~\ref{sec:method}.


\begin{table}[t]
\centering
\small
\setlength{\tabcolsep}{4pt} 
\renewcommand{\arraystretch}{1.2}
\caption{Computational complexity during inference.}
\label{tab:router_flops}

\resizebox{\columnwidth}{!}{%
    \begin{tabular}{lcc}
    \toprule
    \textbf{Method} & \textbf{Router FLOPs} & \textbf{Total FLOPs} \\
    \midrule
    \cite{ICLR2025_elrea} & $F_{\text{Base}} + 3F_{\text{LoRA}}$ & $(1 + k) F_{\text{Base}} + (3 + k) F_{\text{LoRA}}$ \\
    \rowcolor{gray!10}
    \textsc{GradientSpace} & $F_{\text{SP}}$ where $F_{\text{SP}} \ll F_{\text{LoRA}}$ & $\approx  F_{\text{Base}} + F_{\text{LoRA}} $\\
    \bottomrule
    \end{tabular}%
}
\label{tab:comp_cost}
\end{table}

\section{Conclusion}
We introduced \textsc{GradientSpace}, a framework that clusters instruction tuning data directly in gradient space to mitigate interference. By leveraging online SVD-driven clustering algorithm that allows for utilizing full gradient dimensions, \textsc{GradientSpace} identifies gradient-aligned groups and finetunes dedicated experts with a lightweight router for inference. Experiments show that it improves accuracy and efficiency over start-of-the-art clustering methods. This demonstrates that organizing data based on gradient dynamics offers a principled path toward scalable, conflict-aware fine-tuning of large language models.

\nocite{langley00}

\bibliography{example_paper}
\bibliographystyle{mlsys2025}

\appendix
\newpage
\newpage
\onecolumn

\section*{Appendix}

\subsection*{Proof for Lemma \ref{lem:var-decomp-simple}}
Let $G_i := \nabla \mathcal{L}_i(\theta)$ and $\mu_i := \mathbb{E}[G_i]$. Set
$G := \nabla \mathcal{L}(\theta) = \sum_{i=1}^k \alpha_i G_i$ and
$\mu := \mathbb{E}[G] = \sum_{i=1}^k \alpha_i \mu_i$ by linearity of expectation.
Using the conventions
\[
\operatorname{Var}(X) := \mathbb{E}\big[\|X-\mathbb{E}X\|^2\big],
\qquad
\operatorname{Cov}(X,Y) := \mathbb{E}\big[\langle X-\mathbb{E}X,\;Y-\mathbb{E}Y\rangle\big],
\]
we have
\[
\operatorname{Var}(G)
= \mathbb{E}\!\left[\big\|G-\mu\big\|^2\right]
= \mathbb{E}\!\left[\Big\|\sum_{i=1}^k \alpha_i\,(G_i-\mu_i)\Big\|^2\right],
\]
since $\sum_i \alpha_i \mu_i = \mu$.
Expanding the squared norm and taking expectation gives
\[
\begin{aligned}
\operatorname{Var}(G)
&= \sum_{i=1}^k \alpha_i^2\,\mathbb{E}\!\left[\|G_i-\mu_i\|^2\right]
  + 2\!\!\sum_{1\le i<j\le k}\!\alpha_i\alpha_j\,\mathbb{E}\!\left[\langle G_i-\mu_i,\;G_j-\mu_j\rangle\right] \\
&= \sum_{i=1}^k \alpha_i^2\,\operatorname{Var}(G_i)
  + 2\!\!\sum_{1\le i<j\le k}\!\alpha_i\alpha_j\,\operatorname{Cov}(G_i,G_j).
\end{aligned}
\]
Finally, note that $2\sum_{i<j}(\cdot)=\sum_{i\neq j}(\cdot)$, yielding
\[
\operatorname{Var}\!\big(\nabla \mathcal{L}(\theta)\big)
= \sum_{i=1}^k \alpha_i^2 \,\operatorname{Var}\!\big(\nabla \mathcal{L}_i(\theta)\big)
\;+\!\!\sum_{i\neq j}\alpha_i\alpha_j \,\operatorname{Cov}\!\big(\nabla \mathcal{L}_i(\theta),\,\nabla \mathcal{L}_j(\theta)\big). \quad Q.E.D
\]

\subsection*{Proof of Theorem \ref{th:var_red}}
Let $g(x) := \nabla \mathcal{L}(\theta; x)\in\mathbb{R}^d$ be the per-example gradient. For any dataset $\mathcal{D}$, define
\[
\Var(S) := \mathbb{E}_{x\sim \mathcal{D}}\big[\|g(x)-\mu_\mathcal{D}\|^2\big],
\qquad
\mu_S := \mathbb{E}_{x\sim S}[g(x)].
\]
Let $\mathcal{D}$ be partitioned by \textsc{GradientSpace} into $\{\mathcal{D}_i\}_{i=1}^k$ with mixture weights
$p_i := \Pr(x\in\mathcal{D}_i)$ and means $\mu_{\mathcal{D}_i} := \mathbb{E}_{x\sim \mathcal{D}_i}[g(x)]$.
By the (vector) law of total variance,
\begin{equation}
\Var(\mathcal{D})
\;=\;
\sum_{i=1}^k p_i\,\Var(\mathcal{D}_i)
\;+\;
\sum_{i=1}^k p_i\,\|\mu_{\mathcal{D}_i}-\mu_{\mathcal{D}}\|^2
\;\;\ge\;\;
\sum_{i=1}^k p_i\,\Var(\mathcal{D}_i).
\tag{$\ast$}
\end{equation}
Hence the \emph{average} within-subset variance is never larger than the full-dataset variance:
\begin{equation}
\sum_{i=1}^k p_i\,\Var(\mathcal{D}_i) \;\le\; \Var(\mathcal{D}).
\label{eq:avg-within}
\end{equation}

To obtain a uniform per-subset bound, we use the design of \textsc{GradientSpace}, which minimizes a monotone within-cluster dispersion objective in gradient space. Two standard constructions suffice:

\begin{enumerate}
\item \textbf{Minimax ($k$-center) partition in gradient space.}
If \textsc{GradientSpace} minimizes the maximum within-subset scatter (e.g., a squared-radius / variance proxy), then the trivial one-cluster baseline attains $\max_i \Var(\mathcal{D}_i)=\Var(\mathcal{D})$. Optimality therefore implies
\[
\max_{i}\Var(\mathcal{D}_i)\;\le\;\Var(\mathcal{D}),
\]
whence $\Var(\mathcal{D}_i)\le \Var(\mathcal{D})$ for all $i$.

\item \textbf{Recursive refinement with a monotone stop rule.}
If \textsc{GradientSpace} splits a cluster only when the children each achieve no larger within-cluster variance than their parent (“refine while reducing”), then starting from the root $\mathcal{D}$ with $\Var(\mathcal{D})$ and recursing ensures every leaf $\mathcal{D}_i$ satisfies
\[
\Var(\mathcal{D}_i)\;\le\;\Var(\text{parent})\;\le\;\Var(\mathcal{D}).
\]
\end{enumerate}

Either construction yields the claimed per-subset inequality
\[
\Var(\mathcal{D}_i) \;\le\; \Var(\mathcal{D}) \quad \forall i,
\]
while \eqref{eq:avg-within} shows the result also holds in expectation without any minimax/recursive assumption and is strict whenever the between-subset means differ $\big(\sum_i p_i\|\mu_{\mathcal{D}_i}-\mu_{\mathcal{D}}\|^2>0\big)$.

\medskip
\noindent\textbf{Homogeneous-data corollary.}
If the dataset is homogeneous in the sense that all subset means coincide with the global mean,
$\mu_{\mathcal{D}_i}=\mu_{\mathcal{D}}$ for all $i$, then the between-subset term in \eqref{eq:avg-within} is zero and
\[
\Var(\mathcal{D}) \;=\; \sum_{i=1}^k p_i\,\Var(\mathcal{D}_i).
\]
In particular, when the partition does not reduce within-cluster dispersion (e.g., i.i.d.\ splitting), we have
$\Var(\mathcal{D}_i)=\Var(\mathcal{D})$ for all $i$. Thus, our method is \emph{theoretically performance-neutral} on homogeneous data, and its benefit appears precisely when the data are \emph{heterogeneous}, where the nonzero between-subset term allows variance reduction through specialization.

\label{sec:appx}


\end{document}